\documentclass[a4paper, 10pt, conference]{cssconf}        
\pdfoutput=1
\IEEEoverridecommandlockouts                              
\usepackage{graphicx} 
\usepackage{times} 
\usepackage{amsmath} 
\usepackage{amssymb}  
\usepackage{bm}  

\usepackage{multirow,color}
\usepackage{hyperref}
\DeclareMathAlphabet{\bdmath}{OML}{cmm}{b}{it}     
\mathversion{normal}

\def\0{\mathbf{0}}

\def\F{\bdmath{F}}
\def\J{\bdmath{J}}
\def\K{\bdmath{K}}

\def\W{\bdmath{W}}

\def\q{\bdmath{q}}
\def\u{\bdmath{u}}
\def\x{\bdmath{x}}

\def\qdot{\dot{\q}}

\sloppypar

\title{\LARGE \bf
Validation of a Control Algorithm for Human-like Reaching Motion using 7-DOF Arm and 19-DOF Hand-Arm Systems}

\author{Tapomayukh Bhattacharjee, Yonghwan Oh* and Sang-Rok Oh
\thanks{During this work, T. Bhattacharjee was with the Interaction and Robotics Research Center, KIST, 39-1 Hawolgok-dong, Wolsong Gil 5, Seongbuk-gu, Seoul 136-791, South-Korea. Now, he is with Georgia Institute of Technology, USA}
\thanks{Y. Oh, and S. R. Oh are with the Interaction and Robotics Research Center, KIST, 39-1 Hawolgok-dong, Wolsong Gil 5,
    Seongbuk-gu, Seoul 136-791, South-Korea
        {\tt\small \{oyh, sroh\}@kist.re.kr}}%
\thanks{*Y. Oh is the corresponding author}
}

\begin{document}

\maketitle
\thispagestyle{empty}
\pagestyle{empty}

\begin{abstract}
This technical report gives an overview of our work on control algorithms dealing with redundant robot systems for achieving human-like motion characteristics. Previously, we developed a novel control law to exhibit human-motion characteristics in redundant robot arm systems as well as arm-trunk systems for reaching tasks \cite{01}, \cite{02}. This newly developed method nullifies the need for the computation of pseudo-inverse of Jacobian while the formulation and optimization of any artificial performance index is not necessary. The time-varying properties of the muscle stiffness and damping as well as the low-pass filter characteristics of human muscles have been modeled by the proposed control law to generate human-motion characteristics for reaching motion like quasi-straight line trajectory of the end-effector and symmetric bell shaped velocity profile. This report focuses on the experiments performed using a 7-DOF redundant robot-arm system which proved the effectiveness of this algorithm in imitating human-like motion characteristics. In addition, we extended this algorithm to a 19-DOF Hand-Arm System for a reach-to-grasp task. Simulations using the 19-DOF Hand-Arm System show the effectiveness of the proposed scheme for effective human-like hand-arm coordination in reach-to-grasp tasks for pinch and envelope grasps on objects of different shapes such as a box, a cylinder, and a sphere.
\end{abstract}

\newcommand\pkg[1]{\textsf{#1}}
\newcommand\file[1]{\texttt{#1}}

\newcommand{\norm}[1]{\left\lVert#1\right\rVert}


\section{INTRODUCTION}\label{sec:intro}

Robots have increasingly been used in human environments for dexterous manipulation. Redundancy is almost a prerequisite in such cases because it can be used to manipulate or reach objects in human environments dexterously. Control of such redundant humanoid arm systems for reaching motion should achieve human-motion characteristics like quasi-straight line trajectory of human arm and bell-shaped tangential velocity profile for their acceptability in human society \cite{01}, \cite{02}. The authors, in their previous work in \cite{01}, \cite{02} have developed a control algorithm which can imitate the common human-motion characteristics for redundant arm as well as arm-trunk systems. The new control scheme successfully demonstrates both quasi-straight line end-point trajectory and symmetric bell-shaped velocity profile as well as the temporal characteristics of such arm-trunk motion \cite{01}, \cite{02}.

In this short report, we validate the theory using real-time experimental results using a 7-DOF Robotic Arm system which prove the efficacy of this algorithm in presence of non-linear uncertainties. We also extend this algorithm to multiple-DOF Hand-Arm systems for Reach-to-Grasp tasks. We test this extended formulation using a 19-DOF Robotic Hand-Arm System in a real-time simulation software environment called {\it RoboticsLab} for various Reach-to-Grasp tasks which involve grasping objects of different shapes such as a box, a cylinder, and a sphere using pinch and envelope grasps.

The rest of the report is organized as follows. At first, we provide an overview of our previously developed control law based on \cite{01}, \cite{02} in Section \ref{sec:control}. We then describe our experimental setup for the 7-DOF Arm system in Section \ref{sec:expsetup}. Section \ref{sec:exp} describes the experimental results in detail. Our extended control algorithm for hand-arm coordination in reach-to-grasp tasks is presented in Section \ref{sec:controlhandarm}. Section \ref{sec:simhandarm} describes the results of the extended algorithm for different reach-to-grasp tasks using objects of different shapes with pinch and envelope grasps. \textit{The links to the videos of the above experimental and simulation results are provided in the respective sections.} Section \ref{sec:conclusion} concludes the report.

\section{Human Motion Characterictics}\label{sec:control}
Human motion occurs as a result of various muscle movements, their coordination and the brain muscle communication. For reaching or pointing movements, the primary human motion characteristics which have been identified were the quasi-straight line trajectory of the arm and symmetric bell shaped velocity profile as given in \cite{10,11}. However, for reaching movements involving the trunk, additionally researchers have identified some more temporal characteristics such as the peak velocity of the trunk occurs after the peak velocity of the arm which can also be noticed from \cite{11,12}.
Also, it was noticed that the trunk starts moving with or before the arm by around $10$\,ms but continues to move even after the arm motion has ceased for around $100-200$\,ms more\,\cite{11,12}.

Though there have been many works speculating about the probable causes of such characteristics, there have not been any conclusions and till date the mechanism by which the nervous system controls the redundant degrees-of-freedom to accomplish a task is far from being understood\,\cite{7}. However, researchers have claimed that human muscles can be modeled using a spring and a damper with time-varying characteristics\,\cite{13}. It is argued that muscle stiffness drops during movement while muscle damping increases in motion and the variation of these dynamics are in a sinusoidal manner \cite{13,14}. Also, they are joint-dependent and non-linear in nature\,\cite{14}. Hence, there is a need for a time-varying, non-linear, joint-dependent model for muscle stiffness and damping. Although, Arimoto {\it et. al.}\ have claimed to improve the
velocity profile by using a time-varying stiffness model\,\cite{9}, the results were not satisfactory probably because their time-varying model had a
sudden and drastic increase in the beginning of the motion and then remained constant thereafter which is unlike the actual behavior noticed\,\cite{14}. Also, the model in \cite{9} is developed based on an assumption that the actin-myosin interactions which generate the force could be assumed to be a stochastic process and hence, a gamma distribution had been applied. This assumption had no strong evidence. In addition to the above, it has also been identified that human muscles have low-pass filter characteristics and there is a slight delay in brain to muscle communication due to the nervous system path ways which affects the motion pattern \cite{15,16,17}. A joint dependent low-pass filter is, therefore, needed to exhibit the low-pass filter characteristics of human muscles. Also, low-pass filters generate an amount of delay which can account for the delay in brain to muscle communication in humans. The proposed control algorithm has been developed based on these needs identified.

A new control law has thus been developed based on the needs identified for
characterizing the human motion features in reaching movements involving the trunk. The control law is given below in (\ref{eq:1}).
\begin{equation}
    \u =  - \W_{f}\Big[ \K_{V}\qdot + k\F_{mus}\,\J^{T}(\q)\Delta\x \Big],
                                    \label{eq:1}
\end{equation}
where joint damping matrix $\K_{V}$ is given by
\begin{equation}
    \K_{V} = {\rm diag}\Big[ C_{1}^{*}(t)~~ C_{2}^{*}(t)~~ C_{3}^{*}(t)~~ C_{4}^{*}(t) ~~C_{5}^{*}(t)~~ C_{6}^{*}(t)~~ C_{7}^{*}(t) \Big]   \label{eq:2}
\end{equation}
and
\begin{equation}
    C_{i}^{*}(t) = C_{i}\sin\left( \pi\left( \norm{\Delta\x_{0}}
        - \norm{\Delta\x} \right)/2\norm{\Delta\x_{0}} \right). \label{eq:3}
\end{equation}
$k$ is the virtual spring and $\F_{mus}$ is a bijective joint muscle mapping function given by
\begin{equation}
    \F_{mus} = {\rm diag}\Big[ f_{1}^{*}(t)~~ f_{2}^{*}(t)~~ f_{3}^{*}(t)~~ f_{4}^{*}(t) ~~f_{5}^{*}(t)~~ f_{6}^{*}(t)~~ f_{7}^{*}(t) \Big]   \label{eq:4}
\end{equation}
and
\begin{equation}
    f_{i}^{*}(t) = f_{i}\cos\left( \pi\left( \norm{\Delta\x_{0}}
        - \norm{\Delta\x} \right)/2\norm{\Delta\x_{0}} \right),  \label{eq:5}
\end{equation}
where $f_{i}$ denotes the joint muscle stiffness coefficient and
\begin{equation}
\W_{f} = \text{diag}\left[
        \frac{1}{\tau_{i}s + 1} \right],~~\text{for}~i = 1, \ldots, 7\,.
        \label{eq:lpf}
\end{equation}
As given in the control law shown in (\ref{eq:1}), the control input to the
joint actuators is a function of a damping shaping term given by $\K_{V}$ and a bijective joint muscle mapping function given by $\F_{mus}$.

The damping shaping term $\K_{V}$ is assumed to be a diagonal matrix whose
elements are time-varying functions. This matrix shapes the control input based on the joint actuator velocities and is taken to be diagonal by assuming,
without any loss of generality, that velocity coupling is negligible between the different joints. Each of the diagonal elements is modeled by a sinusoidal
function based on the current distance of the end-effector with that of the
target point. As shown in (\ref{eq:3}), $\norm{\Delta\x_{0}}$ represents the
two-norm of the initial distance of the end-effector with that of the target
point while $\norm{\Delta\x}$ represents the current distance of the end-effector with that of the target point and is updated at every
time-instant. The sinusoidal function is so chosen such that the damping
shaping term vanishes at the start of the motion as $\norm{\Delta\x} \to
\norm{\Delta\x_{0}}$, and then gradually increases during motion till it reaches a maximum value. This time-varying nature is modeled in accordance with the trend noticed for muscle damping in reaching movements \cite{14}. The weighting factors $C_{i}$ are tuned for appropriate trajectory and velocity characteristics where $i = 1, \ldots, 7$.

The $\F_{mus}$ function on the other hand, is modeled as a time-varying joint
dependent bijective muscle mapping function. It is bijective because for every
joint actuator there is a unique corresponding one-to-one mapping to a
time-varying weighting variable which forms the diagonal elements of the function matrix whose dimension is given by the number of degrees of freedom of the robot. This function, as shown in (\ref{eq:4}), acts on the virtual work done to reach the target. A virtual spring is attached to the end-effector which literally pulls the robot end-effector towards the target point where $k$ denotes the stiffness of the virtual spring. The spring force generates a virtual work which is required to be done by the joint actuators to move the robot to that desired target position and is given by $k\J^{T}(\q)\Delta\x$. This mechanical work is then mapped to appropriate joint muscle stiffness values by the mapping function given in (\ref{eq:4}). The time-varying nature of each element is modeled by a cosine function based on the current distance of the end-effector with that of the target point. As shown in (\ref{eq:5}), the cosine function is so chosen such that the muscle stiffness value is maximum before the motion starts as $\norm{\Delta\x} \to \norm{\Delta\x_{0}}$, and then gradually decreases during motion. This time-varying nature is modeled in accordance with the trend noticed for muscle stiffness in reaching movements \cite{14}. The joint muscle stiffness coefficients $f_{i}$ are tuned for appropriate trajectory and velocity characteristics where $i = 1, \ldots, 7$.

\begin{figure}
    \begin{center}
        \includegraphics[width=0.9\columnwidth,keepaspectratio,clip]{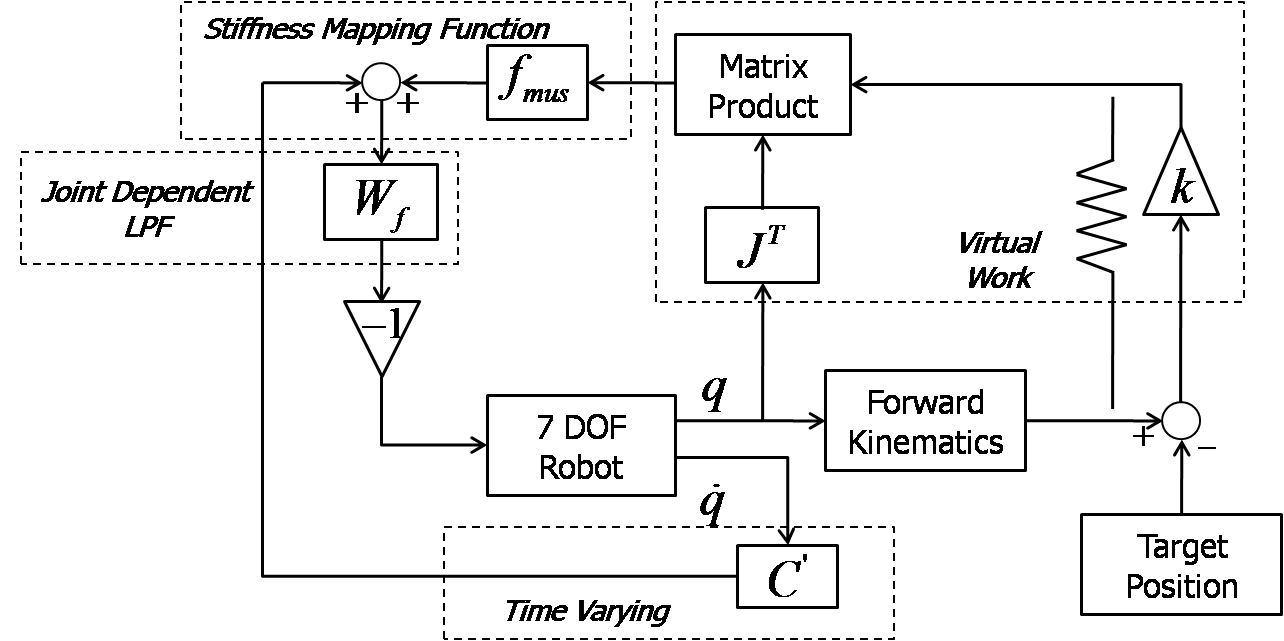}
    \end{center}
    \caption{Block diagram representation of the control law.}  \label{fig:3}
\end{figure}

This control action is then processed by a low-pass filter matrix, as shown in
(\ref{eq:lpf}), because muscles exhibit inherent low-pass filter
characteristics \cite{15,16,17}. The dimension of the matrix corresponds to the number of the degrees of freedom of the robot and each element is modeled as a
first-order low pass filter as shown in (\ref{eq:lpf}) whose time-constant
$\tau_{i}$ is tuned for appropriate behavior. This low-pass filter also
introduces a delay between the control input and the joint actuator torque
output which functions as the brain-muscle communication delay existing in
humans and hence, the time-constant of these joint-dependent filters are
important factors in the design of the control law. The resultant control block diagram is shown in Fig.\,\ref{fig:3}.

\section{Experimental Setup}\label{sec:expsetup}
\begin{figure}
    \centering%
        \includegraphics[width=0.7\columnwidth,keepaspectratio,clip]{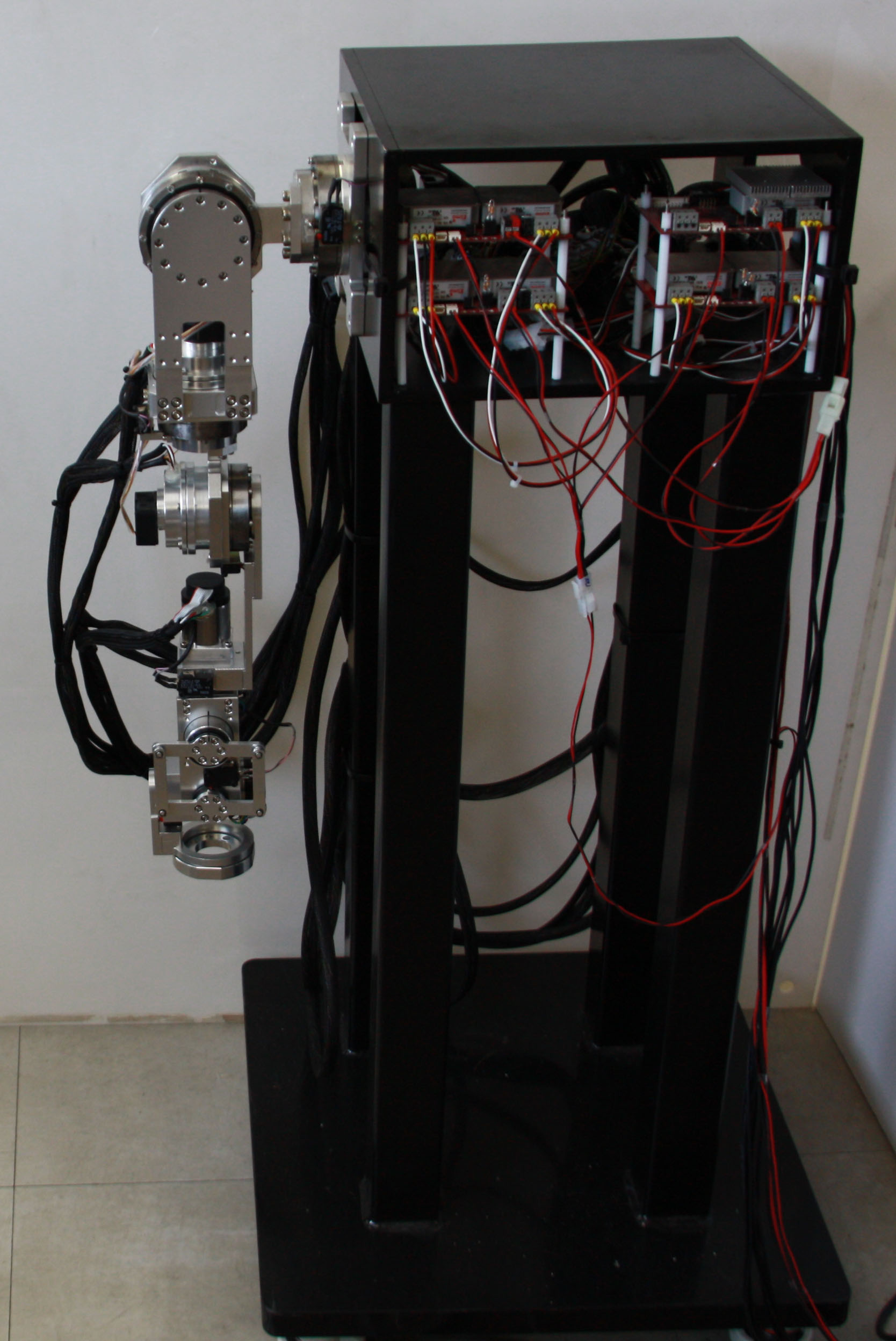}
        \caption{Experimental Setup of a 7-DOF Redundant Robot Arm.}\label{fig:setup}
\end{figure}

To judge the effectiveness of the proposed algorithm, we conducted experiments using a newly built 7-DOF Robot Arm. The experimental setup is as shown in Fig. \ref{fig:setup}.

The first 4 joints are driven by Kollmorgen Motors while the last 3 are driven by Maxon motors. They are connected to the PC through Elmo Digital servo drives and encoders are attached to each of these motors. CANOpen protocol has been used for communication purposes.

Before, we use this 7-DOF robot arm as a platform to verify our developed control law, we need to cancel the undesired dynamic effects and nonlinear uncertainties of this platform. We implemented a Task-Space Disturbance Observer, as given in \cite{03}, to cancel the disturbance terms. For this purpose, we need to compute the Inertia, Coriolis, Gravity, and Friction Terms for compensating the undesired dynamic effects as shown in Section \ref{ssec:dynamics}.

\subsection{Dynamic Formulation of Inertial, Coriolis, Gravity, and Friction Forces}\label{ssec:dynamics}

The computation of the dynamic forces like Inertial, Coriolis, Gravity and Friction forces for a 7-DOF redundant robot is not a trivial task and is computationally very expensive. Hence, we employ the modified recursive Newton-Euler Algorithm for computing these forces \,\cite{13}. Based on \,\cite{13}, our recursive algorithm denoted as $NE_\alpha ^*(q,\dot q,{\dot q_a},\ddot q)$ is given as follows:

\begin{equation}{\omega _i} = {\omega _{i - 1}}\; + \;{\dot q_i}{z_{i - 1}}\end{equation}

\begin{equation}{\omega _{{a_i}}} = {\omega _{{a_{i - 1}}}}\; + \;{\dot q_{{a_i}}}{z_{i - 1}}\end{equation}

\begin{equation}{\dot \omega _i} = {\dot \omega _{i - 1}}\; + \;{\ddot q_i}{z_{i - 1}}\; + \;{\ddot q_i}{\omega _{{a_{i - 1}}}} \times {z_{i - 1}}\end{equation}

\begin{equation}{\ddot O_i} = {\ddot O_{i - 1}}\; + {\dot \omega _i}\; \times {r_{i - 1,i}}\; + \;{\omega _i}\; \times \left( {{\omega _{{a_i}}} \times {r_{i - 1,i}}} \right)\end{equation}

\begin{equation}{\ddot B_i} = {\ddot O_i}\; + {\dot \omega _i}\; \times {r_{i,{B_i}}}\; + \;{\omega _i}\; \times \left( {{\omega _{{a_i}}} \times {r_{i,{B_i}}}} \right)\end{equation}

\begin{equation}{f_i} = {f_{i + 1}}\; + \;{m_i}{\ddot B_i}\end{equation}

\begin{equation}\begin{gathered}{\tau _i} = {\tau _{i + 1}}\; - \;{f_i} \times \left( {{r_{i - 1,i}} + {r_{i,{B_i}}}} \right) + {f_{i + 1}} \times {r_{i,{B_i}}}\;\\
 + {I_i}{\dot \omega _i} + {\omega _{{a_i}}} \times \left( {{I_i}{\omega _i}} \right)\\
 \end{gathered}\end{equation}

\begin{equation}{u_i} = \tau _i^T{z_{i - 1}}\end{equation}

where $\omega_i$ denotes the angular velocity of link $i$, $q_i$ is the angle of joint $i$ and their derivatives have usual meanings, $z_{i}$ is the unit vector in the direction of velocity of joint $i$, $O_i$ is the origin of the $i$-th reference frame, $B_i$ is the origin of the center of mass of link $i$, $f_i$ and $\tau_i$ are the force and torque acting on link $i$, ${\dot q_{{a_i}}}$ is the auxiliary velocity of joint $i$, $r_{i-1,i} = O_i -O_{i-1}$, and $r_{i,B_i} = B_i -O_i$.

Therefore, the joint-space Inertial, Coriolis, and Gravity terms are derived as follows.

\begin{equation}{M_{ij}}(q) = \overset{\lower0.5em\hbox{$\smash{\scriptscriptstyle\frown}$}}{v} _i^TNE_\alpha ^*(q,0,0,{\overset{\lower0.5em\hbox{$\smash{\scriptscriptstyle\frown}$}}{v} _j})\end{equation}

\begin{equation}C_{ij}^*(q,\dot q) = \overset{\lower0.5em\hbox{$\smash{\scriptscriptstyle\frown}$}}{v} _i^TNE_\alpha ^*(q,\dot q,{\overset{\lower0.5em\hbox{$\smash{\scriptscriptstyle\frown}$}}{v} _j},0)\end{equation}

\begin{equation}G_{i}^*(q) = \overset{\lower0.5em\hbox{$\smash{\scriptscriptstyle\frown}$}}{v} _i^TNE_\alpha ^*(q,0,0,0,g)\end{equation}

where $i$, $j$ can be taken from $1$ to the number of joints and ${\overset{\lower0.5em\hbox{$\smash{\scriptscriptstyle\frown}$}}{v} _j}$ denote a column vector with $1$ as the $j$-th term and $0$ as all other terms. $g$ is the acceleration due to gravity and its magnitude is $9.81 m/s^2$.

The Friction term is calculated as,

\begin{equation}f{r_i}\left( q \right) = \eta \dot q + \mu \tanh (\beta \dot q)\end{equation}

where $f{r_i}$ is the joint frictional force, $\eta$ is the coefficient of viscous friction and $\mu$ is the coefficient of coulomb friction.

\subsection{Computation of the Time-Derivative of Jacobian}\label{ssec:jacob}

The Inertial and Coriolis terms, derived in Section \ref{ssec:dynamics}, are joint-space based terms. To convert it to the task-space based formulation so that these can be applied in our task-space disturbance observer, we use the formulation given in \cite{03}. However, we might note that the derivative of the Jacobian needs to be computed to perform this transformation, which is not easy. Here, we explain the manner in which the time-derivative of the Jacobian has been computed for this work.

If the Jacobian can be denoted as given below,

\begin{equation}J = \left[ {\begin{array}{*{20}{c}}
   {{e_i}}  \\
   {{p_i} \times {e_i}}  \\

 \end{array} } \right]\end{equation}

 then,

\begin{equation}\dot J = \left[ {\begin{array}{*{20}{c}}
   {{{\dot e}_i}}  \\
   {\frac{\partial }
{{\partial t}}\left( {{p_i} \times {e_i}} \right)}  \\

 \end{array} } \right] = \left[ {\begin{array}{*{20}{c}}
   {{{\dot e}_i}}  \\
   {{{\dot p}_i} \times {e_i} + {p_i} \times {{\dot e}_i}}  \\

 \end{array} } \right]\end{equation}

 where

\begin{equation}\begin{gathered}
  {{\dot e}_i} = {\omega _i} \times {e_i}\;with\;{\omega _i} = \sum\limits_{j = 1}^{i - 1} {{{\dot q}_j}{e_j}}  \hfill \\
  {{\dot p}_i} = \sum\limits_{k = i}^{j - 1} {{{\dot p}_{k,k + 1}}} \;with\;{{\dot p}_{k,k + 1}} = {\omega _k} \times {p_{k,k + 1}} \hfill \\
\end{gathered} \end{equation}

Thus, the time-derivative of the Jacobian can be computed.

\section{Experimental Results}\label{sec:exp}
After the undesired dynamic effects for the 7-DOF Robot Arm were compensated, we implemented the control algorithm as given in Section \ref{sec:control}. Let~$\bm{I_{i}}$ denote the inertia tensors for all the links of the robot wherein $i = 1, \ldots, 7$ and the diagonal elements are given by $I_{xx},~I_{yy}$, and $I_{zz}$ which are the principal moments of inertia. The non-diagonal elements given by $I_{xy},~I_{yz}$, and $I_{xz}$ are negligible for highly symmetric construction of links. Let $\ell_i$, $COM$, and $m_i$ denote the link lengths, distance of Center of Masses of links from link frames and link masses respectively. The lengths of the links as well as the dynamic parameters such as the link inertia and mass properties are based on manufacturer specifications and rough human measurements, and are given in Table~\ref{tab:1}. The calculated coulomb and viscous friction coefficients of every joint are also given in Table~\ref{tab:1}. All dimensions used in this report are in S.I. units.
\begin{table}
\caption{Robot parameters}  \label{tab:1}
\begin{center}
\begin{tabular}{|c|c|c|c|c|c|c|c|}    \hline
        & \multirow{2}{*}{}Joint & Joint & Joint & Joint & Joint & Joint & Joint\\
        & 1 & 2 & 3 & 4 & 5 & 6 & 7                           \\\hline\hline
    $I_{xx}$    & 0.006 & 0.012 & 0.003 & 0.005 & 0.001 & 0.002 & 0.001 \\\hline
    $I_{yy}$    & 0.001 & 0.002 & 0.000 & 0.001 & 0.000 & 0.001 & 0.000   \\\hline
    $I_{zz}$    & 0.006 & 0.011 & 0.003 & 0.005 & 0.001 & 0.002 & 0.001   \\\hline
    $\ell$     & 0.085 & 0.171 & 0.069 & 0.148 & 0.095 & 0.0 & 0.0755      \\\hline
    $COM$     & 0.011 & 0.091 & 0.007 & 0.051 & 0.058 & 0.0 & 0.0185      \\\hline
      $m$       &  0.81  & 2.096 & 0.538 & 0.407 & 0.459 & 0.396 & 0.135            \\\hline
    $\mu$ & 3.704  & 5.02 & 1.359 & 1.240 & 0.607 & 0.979 & 0.778            \\\hline
    $\eta$ & 1.104  & 2.086 & 1.191 & 1.016 & 0.668 & 0.794 & 0.604            \\\hline
\end{tabular}
\end{center}
\end{table}

Using the above parameters and the mathematical formulation as described in Section \ref{ssec:dynamics}, we compensated for the undesired dynamic effects including friction and gravity effects. $\beta$ for friction compensation was taken as $20$. A video, which shows the results of our friction and gravity compensation methods, is uploaded in \url{https://docs.google.com/leaf?id=0B5IkR-5za8vGanFSMkFwWnJhWHM}. After compensating the undesired dynamics, we implement the developed control law in the 7-DOF robot arm. For implementing the algorithm according to Section \ref{sec:control}, we have selected the parameters as given in Table~\ref{tab:2}. The spring constant of the virtual spring is taken as $k=13$ N/m and the target position was given at $X_{d} = 0.3m, ~Y_{d} = 0.3m, ~and~ Z_{d} = -0.3m$. The initial position of the robot is at $X_{0} = 0.085m, ~Y_{0} = 0.0m, ~and~ Z_{0} = -0.5585m$.
\begin{table}
\caption{Control parameters}  \label{tab:2}
\begin{center}
\begin{tabular}{|c|c|c|c|c|c|c|c|}    \hline
        & \multirow{2}{*}{}Joint & Joint & Joint & Joint & Joint & Joint & Joint\\
        & 1 & 2 & 3 & 4 & 5 & 6 & 7                           \\\hline\hline
    $C_{i}$     & 20 & 10 & 20 & 10 & 10 & 10 & 0.1      \\\hline
    $f_{i}$     & 180 & 40 & 10 & 20 & 1 & 1 & 1      \\\hline
    $\tau_{i}$       &  0.015  & 0.015 & 0.015 & 0.015 & 0.015 & 0.015 & 0.015            \\\hline
\end{tabular}
\end{center}
\end{table}

The results of the experiments are given below in Figs. \ref{fig:exp1} and \ref{fig:exp2} and these results show that the proposed algorithm successfully imitates human-motion characteristics for multi-DOF redundant robotic arm systems for reaching tasks.

\begin{figure}
    \centering%
        \includegraphics[width=0.8\columnwidth,keepaspectratio,clip]{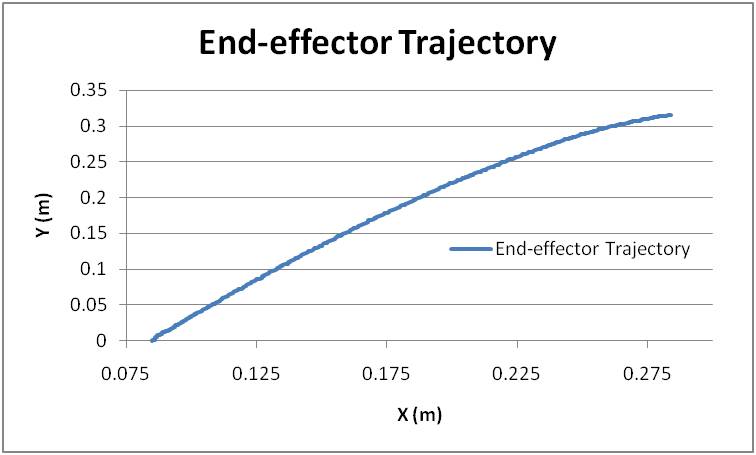}
        \caption{Quasi-Straight Line Trajectory for Reaching Tasks in Non-Gravity Plane.}\label{fig:exp1}
\end{figure}
\begin{figure}
    \centering%
        \includegraphics[width=0.8\columnwidth,keepaspectratio,clip]{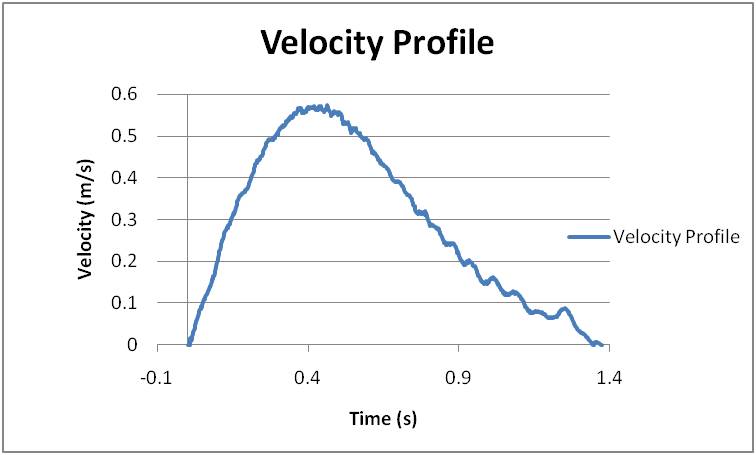}
        \caption{Bell Shaped Velocity Profile for Reaching Tasks.}\label{fig:exp2}
\end{figure}

The experimental video which shows the overall performance of the 7-DOF Robot Arm in reaching to a particular goal with compliance is uploaded in \url{https://docs.google.com/leaf?id=0B5IkR-5za8vGVTlsR2dFWEdIQ2s}.

\section{Extended Hand-Arm Control Law}\label{sec:controlhandarm}
\begin{figure}
    \centering%
        \includegraphics[width=0.7\columnwidth,keepaspectratio,clip]{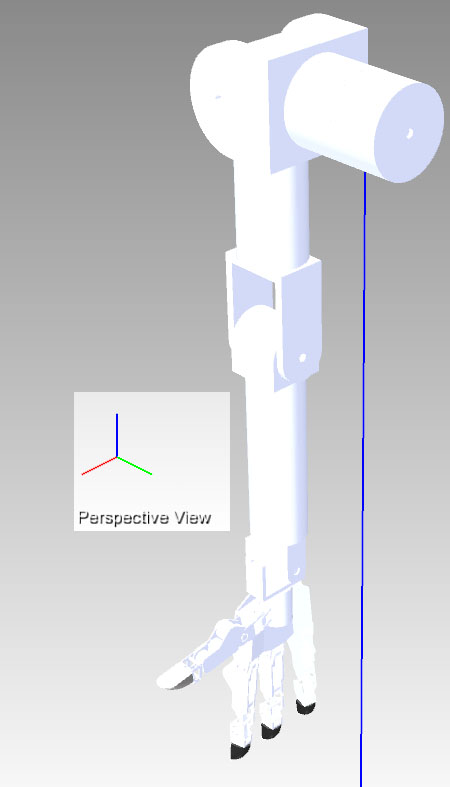}
        \caption{A Robotic Hand-Arm System with 7-DOF Arm and 12-DOF Hand.}\label{fig:hand-arm}
\end{figure}

\begin{figure*}
    \centering%
        \includegraphics[width=1.7\columnwidth,keepaspectratio,clip]{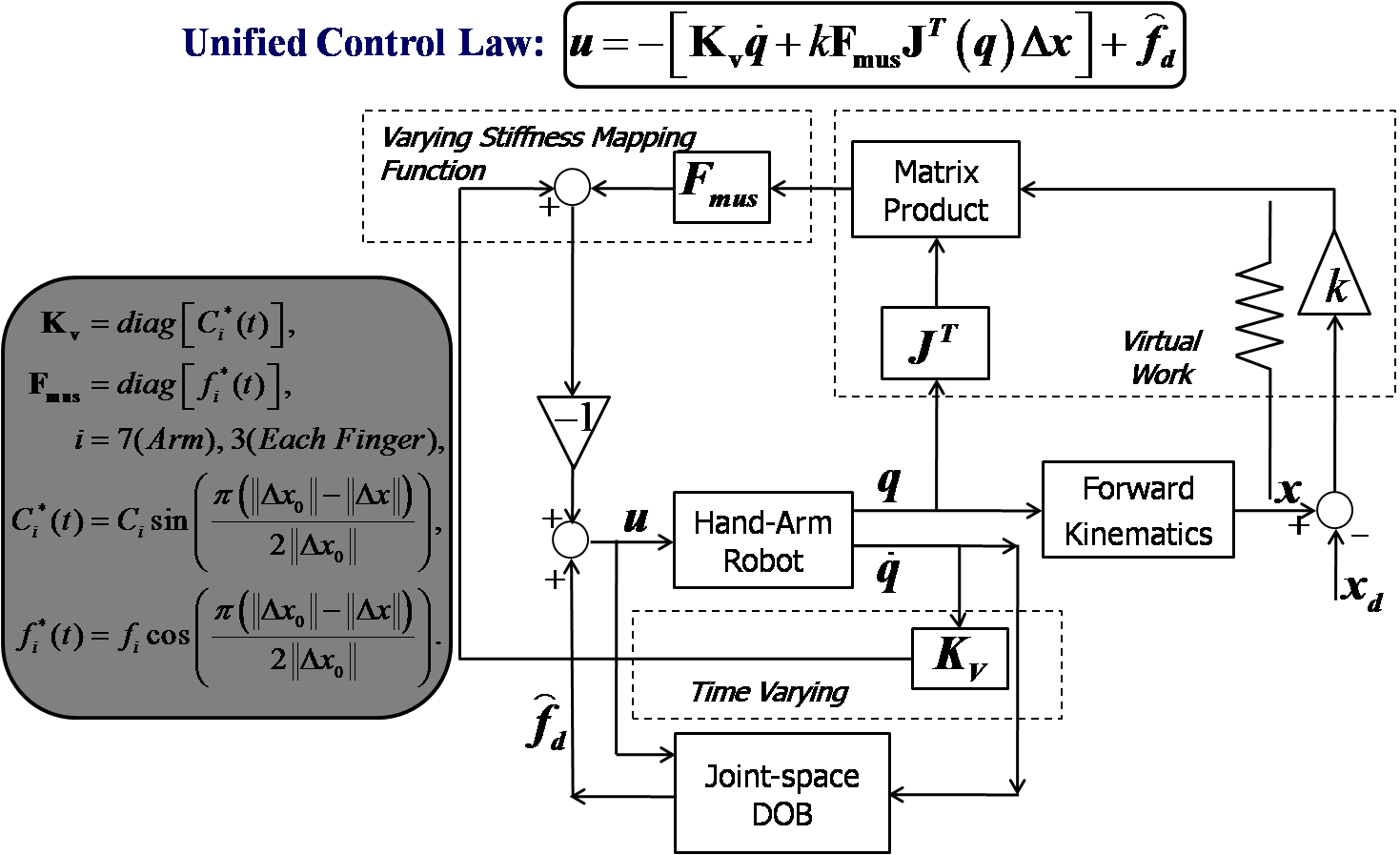}
        \caption{A Block Diagram Representation of the Unified Hand-Arm Control Law for a 7-DOF Arm and 12-DOF Hand using Joint-Space Disturbance Observers.}\label{fig:controlhandarm}
\end{figure*}

We extended the Arm Control Law to a 19-DOF Hand-Arm System to see if the Extended/Unified Hand-Arm Control law can be used for reach-to-grasp tasks. The extension is straight-forward such that it is equivalent to adding 4 serial 3-DOF arms (or fingers) to the 7-DOF robotic arm. However, the main challenge lies in the fact that there is a huge difference in the mechanical parameters such as inertia between the bulky arm and the light fingers and hand. Hence, to cancel the effect of coupling that might effect the finger movements due to the inertia and other dynamic effects of the arm, we employed a joint-space disturbance observer. The entire 19-DOF Hand-Arm system is shown in Fig. \ref{fig:hand-arm}. The Unified Hand-Arm Control Law is shown in Fig. \ref{fig:controlhandarm}.

\section{Hand-Arm Coordination: Simulation Results}\label{sec:simhandarm}
\begin{figure}
    \centering%
        \includegraphics[width=1\columnwidth,keepaspectratio,clip]{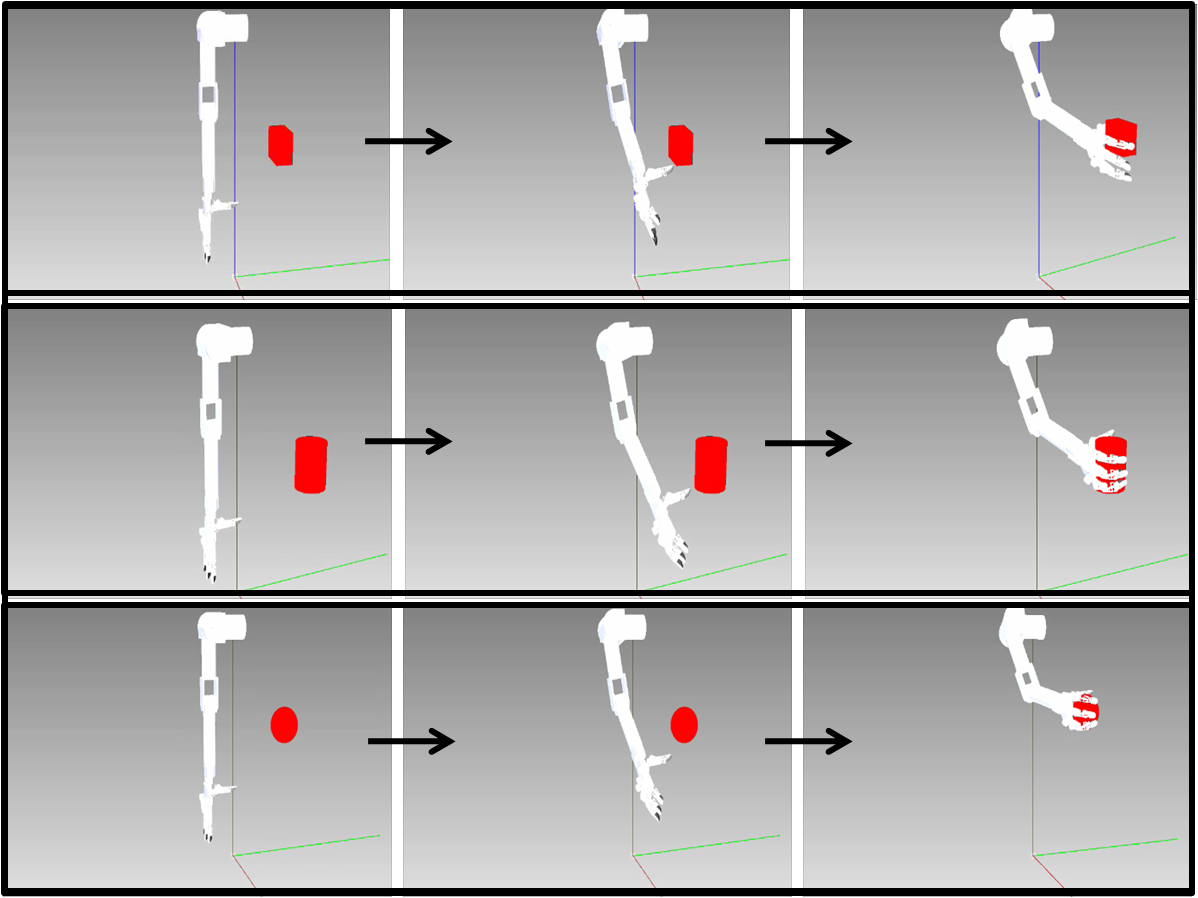}
        \caption{Sequence of steps of a 19-DOF robotic Hand-Arm System grasping three objects using the Unified Hand-Arm Law : \textit{Top: Box with Pinch Grasp}, \textit{Middle: Cylinder with Envelope Grasp}, and \textit{Bottom: Sphere with Pinch Grasp}.}\label{fig:simresult}
\end{figure}

Simulations using the real-time simulation software environment called {\it RoboticsLab} were performed for various Reach-to-Grasp tasks which involve grasping objects of different shapes such as a box, a cylinder, and a sphere using pinch and envelope grasps. Simulation results show the efficacy of the Unified Hand-Arm Control law for such reach-to-grasp tasks for all the above objects and grasp types. The results are shown in Fig. \ref{fig:simresult}.

The resulting simulation video of all the above grasps for all the three different objects is uploaded in \url{https://docs.google.com/leaf?id=0B5IkR-5za8vGZ1N3bkhsTFN3MTQ}.

\section{CONCLUSION}\label{sec:conclusion}

In this report, we presented a novel control algorithm to exhibit the spatial and temporal characteristics of human reaching movements for multi-DOF robot arm systems based on our previous results. The newly developed control algorithm is developed based on the observed human muscle stiffness and damping properties. This control scheme takes into account the time-varying, joint dependent characteristics of the muscle stiffness and damping as well as the low-pass filter characteristics of human muscles and also considers the communication delay in brain-muscle communication. We implemented this control algorithm using a 7-DOF Robotic Arm and the experimental results also show the effectiveness of this algorithm in describing the characteristics of the human reaching motion. We compensated the friction and gravity effects and canceled the undesirable dynamic effects using a Task-Space disturbance Observer. We extended this algorithm to a Unified Hand-Arm Control Law for effective Hand-Arm Coordination in reach-to-grasp tasks. Simulations were conducted using a 19-DOF Hand-Arm system for reaching and grasping objects of different shapes such as a box, a cylinder and a sphere using pinch and envelope grasps. We implemented a joint-space disturbance observer to cancel the coupling effects between the bulky arm with high inertia and the light-weight fingers. Results show the effectiveness of the Unified Hand-Arm Control law for Hand-Arm Coordination in such reach-to-grasp tasks.


\end{document}